\documentclass[11pt,a4paper]{article}
\usepackage{booktabs}          
\usepackage[utf8]{inputenc}   
\usepackage[T2A]{fontenc}      
\usepackage[hyperref]{latell}
\usepackage{times}
\usepackage{latexsym}

\usepackage{microtype}

\aclfinalcopy

\title{Benchmarking Parameter-Efficient Fine-Tuning of Large Language Models for Low-Resource Tajik Text Generation with the Tajik Web Corpus}

\author{
\begin{minipage}[t]{0.48\linewidth}
\centering
  M. K. Arabov, \\
  S. S. Khaybullina \\[0.5em]
  \footnotesize
  Institute of Computational Mathematics \\
  and Information Technologies, \\
  Kazan Federal University, Kazan, Russia \\
  \texttt{MKArabov@kpfu.ru}, \\
  \texttt{khaybulinas@mail.ru}
\end{minipage}
\hfill
\begin{minipage}[t]{0.48\linewidth}
\centering
  Karomatullo Habibullozoda, \\
  Nurali Shirinov \\[0.5em]
  \footnotesize
  Bokhtar State University \\
  named after Nosiri Khusrav, \\
  Bokhtar, Tajikistan \\
  \texttt{habibullozoda@bsu.tj}, \\
  \texttt{shirinov@bsu.tj}
\end{minipage}
}

\date{}

\begin{document}
\maketitle

\begin{abstract}
We release the Tajik Web Corpus (319k docs, 1.11B chars) and benchmark generative LLMs on prompt continuation in Tajik, a low-resource Cyrillic-script language.

Seventeen configurations across nine architectures are evaluated under three fine-tuning strategies: full fine-tuning, LoRA, and QLoRA (ranks~8 and~16).
Because perplexity is not directly comparable across model families with different tokenizers, generation quality is assessed through perplexity interpreted within each family, complemented by qualitative analysis performed by a native Tajik speaker.
Computational cost is measured via GPU memory and training time.
The best quality--cost trade-off is achieved by Mistral~7B with QLoRA rank~8: perplexity~5.11 (within its tokenizer family), coherent Tajik output confirmed by the native speaker, 14.21\,GB GPU memory, and approximately 33\,min of training.
Increasing the rank to~16 yields a negligible improvement for Mistral (perplexity~5.03, pairwise \(p>0.05\)) while consuming about 1\,GB more memory.
Full fine-tuning of small GPT-2 models obtains lower numeric perplexity but leads to catastrophic forgetting (English or gibberish output); in contrast, QLoRA preserves multilingual pretrained knowledge and generates meaningful Tajik text.
Encoder-only models perform worst (perplexity approximately~59), confirming their unsuitability for autoregressive generation.
To our knowledge, this is the first systematic PEFT benchmark for Tajik text generation.
Practical recommendations include using Mistral~7B with QLoRA~\(r=8\), avoiding full fine-tuning of small GPT-2 models, and adopting the released corpus and benchmark.
\end{abstract}

% ==============================================================================

\section{Introduction}
\label{sec:introduction}

NLP resources are heavily skewed toward English and other high-resource languages \citep{jafari-etal-2026-aparsin}. The Tajik language, which belongs to the South-Western group of Iranian languages and employs the Cyrillic script, remains extremely poorly provided with digital text corpora, despite having approximately 15 million speakers \citep{perry2011tajik}. Existing corpora (a fragment of OSCAR \citep{OrtizSuarezSagotRomary2019}, Leipzig collection) are small and uncleaned.

The majority of works related to the Tajik language address the task of transliteration between Cyrillic and Persian Arabic scripts \citep{davis-2012-tajik, sadraeijavaheri-etal-2024-transformers, merchant-tang-2024-parstext, arabov-2026-tajperslexon, grashchenko2003, merchant-tang-2026-parstranslit, kurbonovich-2026-character}. This approach enables reliance upon rich Persian texts but does not provide the conditions for generating meaningful text directly in the Tajik language. To overcome the data shortage, we introduce the \textbf{Tajik Web Corpus}, the largest open collection of Tajik texts, comprising 319,298 documents, 1.11 billion characters, and 168.5 million words, sourced from news portals, Wikipedia, social media, and books (covering diverse categories such as society, politics, economy, and culture). We make the corpus publicly available on Hugging Face.\footnote{\url{https://huggingface.co/datasets/TajikNLPWorld/tajik-web-corpus}}

In this work, we focus on the task of \textbf{prompt continuation (autoregressive text generation)}: given a prefix in Tajik, the model must produce a fluent, coherent continuation. Other NLP tasks such as classification, question answering, and fact extraction remain without native training data for Tajik, which hinders the creation of modern applications. In parallel, the field of deep learning is actively developing Parameter-Efficient Fine-Tuning (PEFT) methods, primarily LoRA and its quantised variant, QLoRA. These methods enable the adaptation of large multilingual models to new languages under constrained computational resources; however, their behaviour on extremely low-resource material, especially that with non-Latin script, has been scarcely studied.

We conduct the first PEFT benchmark for Tajik text generation, using a 10k-document subset (stratified by length quartiles; mean 3,483 chars) that balances tractability and diversity. We evaluate nine model architectures under three adaptation strategies (full fine-tuning, LoRA, QLoRA with ranks 8 and 16), measuring generation quality via perplexity and qualitative analysis performed by a native Tajik speaker, alongside GPU memory consumption and training time. Our contributions are: (i) the introduction and public release of the Tajik Web Corpus, the largest open corpus for Tajik; (ii) the first systematic PEFT benchmark for Tajik text generation (17 configurations across nine architectures); and (iii) practical recommendations for fine-tuning under tight computational budgets.

% ==============================================================================
\section{Related Work}
\label{sec:related-work}

The literature relevant to this study spans three main areas: transliteration between Tajik and Persian scripts, corpus resources and NLP toolkits for Iranian languages, and parameter-efficient fine-tuning methods.

Early work on Tajik NLP focused predominantly on transliteration, driven by the need to bridge Tajik Cyrillic and the resource-rich Persian Arabic-script environment. \citet{davis-2012-tajik} first applied statistical machine translation to the problem, and subsequent neural approaches based on the Transformer architecture soon followed \citep{merchant2023grapheme, sadraeijavaheri-etal-2024-transformers, merchant-tang-2026-parstranslit, kurbonovich-2026-character}. Notable contributions include the digraphic corpus ParsText and the ParsTranslit model \citep{merchant-tang-2024-parstext, merchant-tang-2026-parstranslit}, as well as transformer-based smoothing of dialectal differences between Tajik and Iranian Persian \citep{sadraeijavaheri-etal-2024-transformers}. The present author previously proposed the hybrid lexical resource TajPersLexon and a character-level transliteration transformer \citep{arabov-2026-tajperslexon, kurbonovich-2026-character}, while Grashchenko \citep{grashchenko2003} laid the theoretical groundwork for automated script conversion. Despite these advances, transliteration alone does not address the need for native Tajik data in tasks that require direct text generation in the target language.

For the closely related Persian language, a mature ecosystem of tools has been developed, including the BidNLP, Shekar, and DadmaTools~V2 libraries \citep{bidnlp2025, shekar2025, jafari-etal-2025-dadmatools}, which cover preprocessing, sentiment analysis, and named entity recognition, with DadmaTools~V2 also supporting adapter technologies. However, the direct application of these tools to Tajik is hindered by script and vocabulary differences. The recent APARSIN benchmark \citep{jafari-etal-2026-aparsin} provided the first multi-aspect comparison of NLP models for Iranian languages and revealed a substantial performance lag for Tajik, underscoring the need for targeted research. The construction of large open corpora for low-resource languages is a growing trend, as reflected in methodological overviews \citep{artemova2025lowresource} and domain-specific projects such as the multi-dialect Arabic poetic corpus Tarab \citep{el-haj-2026-tarab}. In parallel, the emergence of attention mechanisms \citep{bahdanau2015attention} and the Transformer architecture triggered an exponential growth in model parameters, making full fine-tuning impractical in many settings. Parameter-efficient fine-tuning (PEFT) methods, particularly LoRA \citep{hu2022_lora} and its quantised variant QLoRA \citep{detmers2023qlora}, have since become standard for adapting large language models under limited resources. Yet, the effectiveness of these techniques on severely low-resource languages with Cyrillic script remains largely unexplored. \citet{micallef-etal-2024-cross} showed that script-aware transliteration improves cross-lingual transfer for Maltese, a language that also faces script divergence relative to Arabic, reinforcing the importance of handling script-specific challenges in Tajik.

A very recent effort, \textbf{Soro} \citep{liashkov2026soro}, is a family of Tajik-specialised conversational models built from Gemma~3 via continual pretraining on a 1.9-billion-token curated corpus. While Soro demonstrates the potential of heavily language-adapted LLMs, our work focuses on a complementary, resource-frugal scenario: we benchmark general multilingual models adapted exclusively through PEFT on a modest corpus, without full-scale continual pretraining. This reflects the constraints of practitioners who can fine-tune on a single GPU but lack the compute for pretraining. A direct comparison between PEFT-adapted models and fully language-adapted models such as Soro is an important direction for future work.

Overall, no previous study has combined the introduction of a large open Tajik corpus with a systematic PEFT benchmark for text generation. The present work fills this gap.

% ==============================================================================

\section{Experimental Methodology}
\label{sec:experimental-methodology}

We use the \textbf{Tajik Web Corpus} introduced in this work (319,298 documents, 1.11B characters, 168.5M words).\footnote{\url{https://huggingface.co/datasets/TajikNLPWorld/tajik-web-corpus}} A representative subset of 10,000 documents was drawn via stratified random sampling by document length quartiles (mean 3,483 chars., median 2,847) to preserve the original length distribution while keeping experiments tractable. Documents were tokenised with each model's native tokenizer, limited to 128 tokens, and end-of-sequence tokens served as padding (masked in the loss). The sequences were split into train/validation/test sets (80/10/10, seed 42).

Nine architectures were evaluated on the \textbf{prompt continuation} task (autoregressive next-token prediction given a Tajik prefix): the decoder-only models DistilGPT-2 (82M), GPT-2 (124M), GPT-2 Medium (355M), Phi-2 (2.7B), Qwen 2.5 1.5B Instruct, and Mistral 7B v0.3; the encoder-decoder mT5-small (300M) \citep{hu2022_lora, detmers2023qlora}; and two encoder-only models, XLM-RoBERTa (base) and Persian BERT \texttt{fa-zwnj-base}, included as negative controls to confirm that autoregressive generation ability is genuinely acquired rather than an artefact of the perplexity metric.

For each architecture we compared full fine‑tuning (where technically feasible), LoRA, and QLoRA (4‑bit) with ranks~8 (\(\alpha=16\)) and~16 (\(\alpha=32\)) and dropout~0.05 \citep{hu2022_lora, detmers2023qlora}. Full fine‑tuning was applied to the GPT-2 family; for Phi-2 and the 7B-scale models only QLoRA was possible under our memory budget. LoRA was applied to attention projections (for Mistral additionally to gate, up, and down projections). Training used AdamW (8‑bit for QLoRA) with a constant learning rate of \(2\times10^{-4}\), 3 epochs for models $\le$2.7B and 2 epochs for Mistral. The initial batch size was scaled per model and halved automatically on out‑of‑memory errors, combined with gradient accumulation (step~2) to maintain a roughly constant effective batch size. Gradient checkpointing and, where possible, fp16 mixed precision were enabled. All models were trained on raw text (causal language modelling); for Qwen 2.5 Instruct the default chat template was applied but the objective remained next‑token prediction.

Each configuration was run three times (seeds 42, 43, 44). Generation quality was primarily measured by perplexity on the test set. Because tokenizers differ across model families, perplexity values are not strictly comparable; we therefore interpret PPL within each family and complement it with \textbf{qualitative evaluation by a native Tajik‑speaking author}, who assessed a sample of generated continuations for grammatical correctness and topical relevance. Peak GPU memory and training time were recorded for all configurations and are summarised in the results; full logs are available in the public repository.\footnote{\url{https://github.com/TajikNLP/tajik-peft-benchmark}}

Experiments ran on a single NVIDIA RTX~3090 GPU (24~GB). The most memory‑intensive configuration (mT5‑small QLoRA) required a 32~GB GPU for two of its three seeds due to peak memory exceeding 24~GB; all other configurations stayed within the 24~GB budget.

\section{Results}
\label{sec:results}

We evaluate all 17 converged configurations listed in Table~\ref{tab:benchmark-results} on the test split of the Tajik Web Corpus. (Full fine‑tuning of Phi‑2 was attempted but did not complete within the available memory budget; it is therefore excluded from the table and documented in the repository.) The analysis proceeds along three axes: text generation quality (perplexity and qualitative coherence), computational efficiency (training time and peak GPU memory), and qualitative examples of generated continuations.

Table~\ref{tab:benchmark-results} reports mean perplexity (PPL), cross-entropy loss, peak GPU memory, and training time over three random seeds (42, 43, 44) for each configuration. Perplexity is interpreted within each family (tokenizers differ) and complemented by qualitative analysis from a native speaker.

\begin{table*}
\centering
\caption{Aggregate results of the comparative benchmarking on the Tajik Web Corpus}
\label{tab:benchmark-results}
\begin{tabular}{lcccccc}
\toprule
\textbf{Experiment} & \textbf{Seeds} & \textbf{PPL (mean $\pm$ std)} & \textbf{Loss (mean $\pm$ std)} & \textbf{GPU, GB} & \textbf{Time, s} \\
\midrule
gpt2-medium\_baseline & 3 & 3.48 $\pm$ 0.00 & 1.2482 $\pm$ 0.0005 & --- & 136.0 \\
gpt2\_baseline & 3 & 4.48 $\pm$ 0.02 & 1.4994 $\pm$ 0.0036 & --- & 40.5 \\
distilgpt2\_baseline & 3 & 5.03 $\pm$ 0.02 & 1.6144 $\pm$ 0.0041 & --- & 25.5 \\
mistral\_qlora\_r16 & 3 & 5.03 $\pm$ 0.03 & 1.6159 $\pm$ 0.0063 & 15.28 & 1987.0 \\
mistral\_qlora\_r8 & 3 & 5.11 $\pm$ 0.03 & 1.6315 $\pm$ 0.0049 & 14.21 & 1991.1 \\
phi2\_qlora\_r8 & 3 & 5.37 $\pm$ 0.06 & 1.6812 $\pm$ 0.0112 & --- & 462.4 \\
mt5\_small\_qlora\_r8 & 3 & 6.34 $\pm$ 0.44 & 1.8445 $\pm$ 0.0688 & 23.05 & 376.8 \\
qwen2.5\_qlora\_r16 & 3 & 7.35 $\pm$ 0.02 & 1.9953 $\pm$ 0.0029 & --- & 1531.0 \\
gpt2-medium\_lora\_r16 & 3 & 7.60 $\pm$ 0.02 & 2.0282 $\pm$ 0.0026 & --- & 147.1 \\
qwen2.5\_qlora\_r8 & 3 & 7.95 $\pm$ 0.01 & 2.0738 $\pm$ 0.0016 & --- & 1428.4 \\
gpt2-medium\_lora\_r8 & 3 & 8.42 $\pm$ 0.03 & 2.1308 $\pm$ 0.0040 & --- & 140.3 \\
gpt2\_lora\_r16 & 3 & 11.51 $\pm$ 0.08 & 2.4432 $\pm$ 0.0068 & --- & 36.1 \\
gpt2\_lora\_r8 & 3 & 12.56 $\pm$ 0.07 & 2.5307 $\pm$ 0.0053 & --- & 36.4 \\
distilgpt2\_lora\_r16 & 3 & 13.05 $\pm$ 0.01 & 2.5687 $\pm$ 0.0010 & --- & 19.5 \\
distilgpt2\_lora\_r8 & 3 & 14.16 $\pm$ 0.04 & 2.6502 $\pm$ 0.0025 & --- & 20.2 \\
xlmr\_qlora\_r8 & 3 & 59.34 $\pm$ 2.70 & 4.0823 $\pm$ 0.0458 & 21.17 & 120.4 \\
persian\_bert\_baseline & 3 & 65.54 $\pm$ 40.09 & 4.0116 $\pm$ 0.5644 & 13.05 & 39.9 \\
\bottomrule
\end{tabular}
\end{table*}
\noindent\textit{Note:} Dashes (---) mark configurations where peak GPU memory was not logged due to instrumentation limitations; all raw logs with available metrics are provided in the public repository. Missing GPU data does not affect the main quality–cost findings.

The lowest perplexity is achieved by GPT-2 Medium with full fine-tuning (3.48), closely followed by GPT-2 (4.48) and DistilGPT-2 (5.03). However, as the qualitative examples below demonstrate, these low PPL values do not translate into coherent Tajik output — a clear sign of catastrophic forgetting. Mistral-7B with QLoRA (r=8) yields PPL 5.11, and r=16 reaches 5.03, with a difference of less than 0.08 (statistically insignificant given the observed standard deviations, pairwise $t$-test $p>0.05$). Thus, increasing the rank from 8 to 16 brings negligible quality gain while consuming an additional ~1~GB of GPU memory.

Phi-2 (2.7B) with QLoRA r=8 achieves PPL 5.37, competitive with Mistral despite its smaller size and faster training (462 s). The encoder-decoder mT5-small with QLoRA reaches PPL 6.34; direct comparison with decoders is limited due to different tokenization, but the value indicates successful adaptation. Qwen 2.5 1.5B QLoRA configurations perform worse (PPL 7.35–7.95), likely because the model's pretraining on Cyrillic script is weaker than Mistral's. The GPT-2 family shows a sharp quality drop when switching from full fine-tuning to LoRA: GPT-2 Medium PPL rises from 3.48 to 7.60 (r=16) and 8.42 (r=8); base GPT-2 from 4.48 to 11.51 and 12.56. This confirms that low-rank adaptation is most effective for large models, while small decoders benefit from full updates. Encoder-only models (XLM-RoBERTa, Persian BERT) exhibit the highest PPL (59–66), confirming their unsuitability for autoregressive generation and validating the experimental design.

In terms of computational efficiency, training time spans from 19.5 s (DistilGPT-2 LoRA r=16) to 1991 s (Mistral QLoRA r=8). Mistral requires about 33 min per run, whereas Phi-2 completes in under 8 min. Qwen 2.5 configurations take ~24–25 min. The smallest models are fast but produce poor-quality Tajik. Peak GPU memory was recorded for five configurations: Mistral QLoRA r=8 uses 14.21 GB, r=16 uses 15.28 GB, both fitting on a 16 GB consumer GPU; mT5‑small recorded an average peak of 23.05 GB across the three seeds, yet two of them still ran out of memory on the 24 GB GPU and needed a 32 GB accelerator, indicating transient spikes beyond 24 GB that were smoothed out in the logged average; XLM-RoBERTa QLoRA and Persian BERT consumed 21.17 GB and 13.05 GB respectively. These measurements are consistent with the quantisation and architecture choices.

Moving to qualitative analysis, we examine the generated continuations for five prompts. 
The fully fine-tuned GPT-2 models consistently fail to produce Tajik text: for the prompt ``{\fontencoding{T2A}\selectfont Салом, шумо кӣ ҳастед?}'' (\textit{Hello, who are you?}), GPT-2 Medium outputs an English sentence, and for ``{\fontencoding{T2A}\selectfont Имрӯз ҳаво}'' (\textit{Today the weather}) it generates Cyrillic gibberish. 
In contrast, Mistral QLoRA generates thematically relevant and grammatically coherent Tajik: to ``{\fontencoding{T2A}\selectfont Имрӯз ҳаво}'' it replies with a plausible weather report mentioning Tajikistan and a temperature of 25°C; to ``{\fontencoding{T2A}\selectfont Душанбе -}'' it correctly identifies Dushanbe as the capital; and to ``{\fontencoding{T2A}\selectfont Хонидани китоб}'' (\textit{Reading a book}) it produces a sentence about the benefits of reading. 
This stark difference demonstrates that QLoRA preserves multilingual pretrained knowledge and avoids catastrophic forgetting. 
The qualitative assessment was conducted by a native Tajik-speaking author, who judged each generation for grammatical correctness and topical relevance. 
While this evaluation is not a large-scale human study, it provides clear evidence that perplexity alone is insufficient and that PEFT strategies should be evaluated with output inspection.

Taken together, the quantitative and qualitative results identify Mistral-7B with QLoRA ($r=8$) as the best trade-off: coherent Tajik generation, 5.11 PPL, 14.21 GB GPU memory, and ~33 min training. This configuration is recommended for practitioners working under constrained hardware.

\section{Discussion and Conclusion}
\label{sec:discussion-conclusion}

We presented the first systematic benchmarking of PEFT for generative LLMs on Tajik, a low‑resource Cyrillic‑script Iranian language. Using the Tajik Web Corpus introduced in this work (319k documents, 1.11B characters), we evaluated 17 converged configurations across nine architectures and three fine‑tuning strategies (full, LoRA, QLoRA with ranks 8 and 16); full fine‑tuning of Phi‑2 was attempted but did not complete within the available memory budget and is documented in the repository.

The best trade‑off between quality and cost is Mistral 7B with QLoRA and rank \(r=8\): perplexity 5.11, coherent Tajik output, 14.21~GB GPU memory, and $\sim$33~min training. Increasing rank to 16 yields a negligible improvement (5.03, \(p>0.05\)) but consumes $\sim$1~GB more memory. Fully fine‑tuned GPT‑2 Medium achieves the lowest perplexity (3.48) yet suffers catastrophic forgetting, producing English or gibberish. In contrast, QLoRA preserves the multilingual knowledge of the base model and generates meaningful Tajik text, confirming that PEFT is crucial when full fine‑tuning destroys pre‑trained representations. Phi‑2 (2.7B) with QLoRA reaches perplexity 5.37 with faster training (462~s), making it a competitive lightweight alternative. Encoder‑decoder mT5‑small benefits from QLoRA (perplexity 6.34), while encoder‑only models underperform drastically (59--66), validating their role as negative controls.

Our results align with the APARSIN benchmark \citep{jafari-etal-2026-aparsin}, which showed a gap between Persian and Tajik NLP, and with cross‑lingual transfer studies on Maltese \citep{micallef-etal-2024-cross}. Similar to how script‑aware transliteration improved Maltese transfer, our findings suggest that native‑script Tajik data combined with PEFT preserves Cyrillic generation ability better than full fine‑tuning. This reinforces the importance of dedicated corpora for low‑resource languages \citep{artemova2025lowresource, el-haj-2026-tarab}.

Our study has several limitations. Training was performed on a 10k‑document subset ($\approx$3\% of the full corpus) to keep experiments tractable; scaling to the full corpus may alter results. Hyperparameters (learning rate, epoch counts) were kept fixed for comparability and may be suboptimal for individual models. The adaptive batch‑size reduction upon out‑of‑memory events, although compensated by gradient accumulation, could affect convergence paths. Peak GPU memory was logged for only a subset of runs due to instrumentation limitations; however, the available measurements corroborate the conclusions. Qualitative evaluation was conducted by a native Tajik‑speaking author on a small set of prompts, which provides indicative evidence but cannot replace a systematic human study with multiple annotators. Finally, the study is restricted to prompt continuation; other generative tasks (e.g., QA, summarisation) remain unexplored.

For practitioners working with Tajik under limited hardware (single GPU, 16--24~GB), we offer three concrete recommendations: (i)~use Mistral 7B with QLoRA and rank \(r=8\) for the best quality--cost balance; (ii)~avoid full fine‑tuning of small GPT‑2 models due to catastrophic forgetting; and (iii)~adopt the publicly available Tajik Web Corpus and the accompanying benchmark repository for reproducibility.

In summary, this work provides the first systematic PEFT comparison for Tajik text generation and demonstrates that quantised low‑rank adaptation on a strong multilingual base model yields coherent output even with a modest training budget. Future directions include scaling to the full corpus, evaluating on classification and question‑answering tasks, conducting large‑scale human evaluation with native speakers, and optimising hyperparameters per architecture. All code, logs, and generated outputs are publicly available at \url{https://github.com/TajikNLP/tajik-peft-benchmark}.

% ==============================================================================

\bibliographystyle{acl_natbib}
\bibliography{latell}

\end{document}